\documentclass[sn-mathphys-num]{sn-jnl}

\usepackage{graphicx}%
\usepackage{multirow}%
\usepackage{amsmath,amssymb,amsfonts}%
\usepackage{amsthm}%
\usepackage{mathrsfs}%
\usepackage[title]{appendix}%
\usepackage{xurl}%
\usepackage{textcomp}%
\usepackage{manyfoot}%
\usepackage{booktabs}%
\usepackage{algorithm}%
\usepackage{algorithmicx}%
\usepackage{algpseudocode}%
\usepackage{listings}%
\usepackage{subfigure}
\usepackage{array}
\usepackage{adjustbox}
\usepackage{tabularx}
\usepackage{booktabs}      
\usepackage{hyperref}
\usepackage{breakurl}

\theoremstyle{thmstyleone}%
\theoremstyle{thmstyletwo}%

\theoremstyle{thmstylethree}%

\begin{document}

\title[Article Title]{Event-Based Method for High-Speed 3D Deformation Measurement under Extreme Illumination Conditions}


\author[1,2]{\fnm{Banglei} \sur{Guan}}

\author*[1,2]{\fnm{Yifei} \sur{Bian}}\email{bianyifei18@nudt.edu.cn}

\author[1,2]{\fnm{Zibin} \sur{Liu}}

\author[1,2]{\fnm{Haoyang} \sur{Li}}

\author[1,2]{\fnm{Xuanyu} \sur{Bai}}

\author[1,2]{\fnm{Taihang} \sur{Lei}}

\author[1,2]{\fnm{Bin} \sur{Li}}

\author[1,2]{\fnm{Yang} \sur{Shang}}

\author[1,2]{\fnm{Qifeng} \sur{Yu}}

\affil[1]{College of Aerospace Science and Engineering, National University of Defense Technology, Changsha 410073, China}

\affil[2]{Hunan Provincial Key Laboratory of Image Measurement and Vision Navigation, Changsha 410073, China}

\abstract{
\noindent $\textbf{Background}$ Large engineering structures, such as space launch towers and suspension bridges, are subjected to extreme forces that cause high-speed 3D deformation and compromise safety. These structures typically operate under extreme illumination conditions. Traditional cameras often struggle to handle strong light intensity, leading to overexposure due to their limited dynamic range.

\noindent $\textbf{Objective}$ Event cameras have emerged as a compelling alternative to traditional cameras in high dynamic range and low-latency applications. This paper presents an integrated method, from calibration to measurement, using a multi-event camera array for high-speed 3D deformation monitoring of structures in extreme illumination conditions.

\noindent $\textbf{Methods}$ Firstly, the proposed method combines the characteristics of the asynchronous event stream and temporal correlation analysis to extract the corresponding marker center point. Subsequently, the method achieves rapid calibration by solving the Kruppa equations in conjunction with a parameter optimization framework. Finally, by employing a unified coordinate transformation and linear intersection, the method enables the measurement of 3D deformation of the target structure.

\noindent $\textbf{Results}$ Experiments confirmed that the relative measurement error is below 0.08$\%$. Field experiments under extreme illumination conditions, including self-calibration of a multi-event camera array and 3D deformation measurement, verified the performance of the proposed method.

\noindent $\textbf{Conclusions}$ This paper addressed the critical limitation of traditional cameras in measuring high-speed 3D deformations under extreme illumination conditions. The experimental results demonstrate that, compared to other methods, the proposed method can accurately measure 3D deformations of structures under harsh lighting conditions, and the relative error of the measured deformation is less than 0.1$\%$.}


\maketitle

\section{Introduction}\label{sec1}
Large engineering structures, such as space launch towers~\cite{bib44} and suspension bridges~\cite{bib37,bib38,bib41}, exhibit high-speed 3D deformations under extreme mechanical loading, which compromise structural safety. The structures typically operate under extreme illumination conditions\cite{bib39,bib40,bib42}. Camera-based optical measurement methods are widely employed in structural health monitoring and various other fields~\cite{bib31}. Nevertheless, traditional cameras are susceptible to overexposure and related issues under extreme lighting conditions. These issues are attributable to hardware sensor limitations. As a result, the precision of structural deformation measurements may be compromised.

More recently, event cameras have been explored as a potential sensor to address the issues above. The event camera is an innovative, dynamic vision sensor inspired by biological visual mechanisms~\cite{bib15,bib43}. In contrast to traditional cameras that capture images at a fixed frame rate, event cameras feature pixels that respond asynchronously to changes in logarithmic luminosity~\cite{bib24,bib20}. When the absolute difference between the current logarithmic intensity and the one at the time of the most recent event exceeds a predefined threshold, the "event" is triggered. Event cameras offer several advantages, including a high dynamic range ($\sim 140 \text{dB}$), high temporal resolution, low latency ($\sim 1\mu \text{s}$), and very low power consumption, making them suitable for deployment in real-time field-based photogrammetry~\cite{bib21,bib19}.
\begin{figure*}[htbp]
\centering\includegraphics[width=\linewidth]{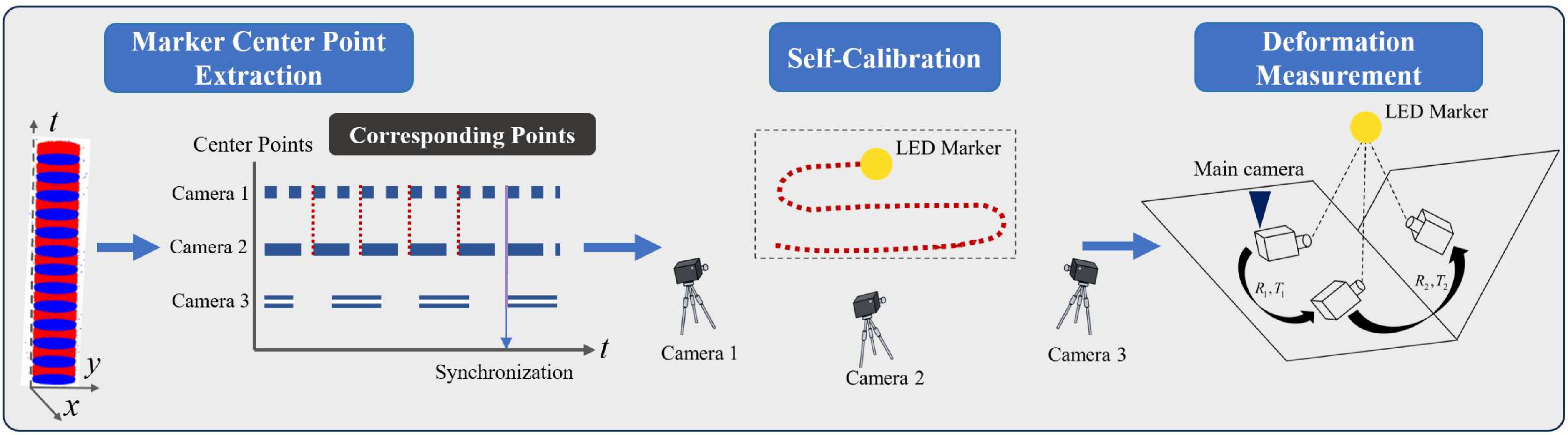}
\caption{Flow diagram of the event-based measurement method for high-speed 3D deformation under extreme illumination conditions.}
\label{fig1}
\end{figure*}

Currently, studies using event cameras for measuring structural deformation have yielded promising results. Structural deformation optical measurement based on asynchronous sparse event stream can be mainly divided into the following three forms: (1) global deformation measurement based on event stream. Lai et al.~\cite{bib32} utilized event cameras to measure structural deformation and proposed a new sparse identification framework based on physical information, enabling vibration monitoring and mechanical analysis of beams. This method requires prior knowledge of the potential dynamics of the monitored structure and is currently only applicable to simpler structures that do not utilize cooperative markers. It only uses event cameras to monitor structural deformation and is not suitable when the ambient light brightness is low. (2) Deformation measurement of structural key nodes based on event stream. Lv et al.~\cite{bib33} proposed a structural plane vibration frequency detection method based on event cameras and developed two algorithms for vibration frequency detection using event stream: tag tracking and event counting, achieving high-precision measurement of vibration frequencies in the range of 10-190 Hz. However, the measurement results are related to the amplitude. As the amplitude decreases, the frequency measurement error increases. When the amplitude is less than 3 pixels, the frequency measurement error exceeds 30$\%$, making the measurement results unreliable. However, to achieve 3D deformation measurement of structural key points,the first step is to extract the corresponding marker centers from asynchronous event streams. (3) Fusion measurement of event stream and other optical measurement methods. Zhu et al.~\cite{bib36} combined event cameras with digital speckle technology to observe blinking markers through event cameras, and then measured the strain of the structure using digital image correlation (DIC), achieving high-speed deformation measurement. However, this method requires spraying the marker on the surface of the structure, and the quality of the speckle affects the error of DIC~\cite{bib45,bib46}. Cheng et al.~\cite{CHENG2018436,CHENG2019314,CHENG2021106286} introduced a novel method based on CT imaging for measuring structural deformation. Compared to event cameras that can measure macroscopic structural deformation, their approach enables the analysis of particle-scale kinematics within materials, which is crucial to facilitate the development and validation of advanced discrete element models for crushable soils that incorporate realistic particle shapes. In summary, existing methods for measuring 3D structural deformation are currently limited.

To apply event cameras to applications such as the aforementioned photogrammetric tasks, precise calibration of the multi-event camera system constitutes an essential prerequisite~\cite{bib7,bib13,bib12}. Calibration methods for traditional cameras can be classified into two categories: photogrammetric calibration and self-calibration~\cite{bib4,bib6,bib29,bib47}. Photogrammetric calibration utilizes a precise calibration object with a known 3D geometry. It establishes geometric correspondences between 3D markers and their two-dimensional projections to calculate the intrinsics and extrinsics~\cite{bib4,bib7}. Such methods require expensive calibration instrumentation and an elaborate setup~\cite{bib10}. By moving a camera within a static environment, if the same camera takes images with fixed intrinsics, correspondences across three frames are sufficient to ascertain both intrinsics and extrinsics~\cite{bib5,bib8,bib9}. Nevertheless, the inherent characteristics of multi-event camera systems pose challenges in feature extraction and temporal synchronization~\cite{bib13}. Consequently, traditional camera calibration methods are not directly suitable for event cameras. The commonly used event camera calibration methods currently include motion-driven calibration methods~\cite{bib3,bib2,bib11,bib48} and methods utilizing dynamic illumination objects~\cite{bib1,bib14,bib28}. Motion-driven calibration methods confront challenges. The movement of the calibration object inevitably leads to the original events being sparse and messy. This compromises calibration precision. Meanwhile, in field environments, the use of large calibration board, blinking screens, or pre defined flashing patterns presents practical challenges related to portability and operational flexibility. Meanwhile, calibrating event cameras via custom-fabricated, bulky, dynamic illumination object apparatuses proves cumbersome. 

To solve the aforementioned issues, this paper presents an integrated method, from calibration to measurement, using a multi-event camera array for high-speed 3D deformation monitoring of structures in harsh outdoor lighting environments. The flow diagram of the paper is shown in Fig.~\ref{fig1}. The blinking light emitting diode (LED) marker moves quickly across the fields of view, enabling accurate and fast extraction from the asynchronous event stream. A temporal correlation analysis framework is developed to identify synchronized marker positions across devices. Corresponding points from multiple event cameras have been successfully extracted without additional constraints. Moreover, calibration parameters are found by solving Kruppa constraint equations, while parameter optimization minimizes reprojection errors. The method yields rapid calibration for multi-camera arrays. Ultimately, our approach enables high-speed 3D deformation measurement through the robust integration of multiple event cameras.
The contributions of this paper include:
\begin{itemize}
    \item We propose an integrated method—from calibration to measurement—using a multi-event camera array, in which a LED marker serves both for calibrating the camera array and for measuring structural deformation, thereby enabling high-speed 3D deformation monitoring under extreme illumination conditions.
    \item We propose a marker center extraction method suitable for both camera calibration and high-speed 3D structural deformation measurement. By leveraging asynchronous event streams and spatiotemporal correlation analysis, the method enables accurate corresponding point extraction from event data captured by multiple cameras without additional constraints.   
    \item We propose a fast self-calibration method for a multi-event camera system by solving the Kruppa constraint equations in conjunction with a parameter optimization framework that minimizes reprojection errors. In contrast to traditional methods, it achieves rapid calibration in field environments.
\end{itemize}

The rest of this paper is organized as follows. Section \ref{sec2} presents the proposed methods. Section \ref{sec3} then demonstrates the detailed experimental results and discussion. The paper concludes in Section \ref{sec4}.

\section{Method}\label{sec2}
This section presents a comprehensive method for high-speed 3D deformation measurement in extreme illumination conditions—from camera calibration to measurement. Section \ref{sec2.1} introduces a LED marker design suitable for both event camera array calibration and deformation measuring, along with its extraction method. Building on marker extraction, Section \ref{sec2.2} describes a fast calibration method for the event camera array. Once camera intrinsics and extrinsics are obtained, Section \ref{sec2.3} outlines specific adaptations for marker extraction in deformation measurement and subsequently presents the method for structural deformation measurement.

\subsection{Marker Center Point Extraction}\label{sec2.1}
The distribution of the event clusters $E$ in the spatiotemporal domain can be approximately modeled as a two-dimensional Gaussian distribution $N(\mu,\Sigma)$, characterized by the following probability density function as Eq. (\ref{eq1}):
\begin{equation}
f(\mathbf{x})=\frac{1}{2\pi \centerdot {{\left| \Sigma  \right|}^{{1}/{2}\;}}}\exp \left[ -{1}/{2}\;{{(\mathbf{x}-\mu )}^{T}}{{\Sigma }^{-1}}(\mathbf{x}-\mu ) \right],\mathbf{x}=(x,y),
\label{eq1}
\end{equation}
where the Gaussian mean $\mu$ describes the location of the event clusters and can be replaced with its estimated value as Eq. (\ref{eq2}):
\begin{equation}
\mu =\overline{\mathbf{x}}=\frac{1}{n}\sum\limits_{i=1}^{n}{{{\mathbf{x}}_{i}}},
\label{eq2}
\end{equation}
where $n$ represents the total number of events in a cluster. The covariance matrix is a second-order square matrix that describes the shape and size of the event cluster, as expressed in Eq. (\ref{eq3}):
\begin{equation}
\Sigma =\left[ \begin{matrix}
   \sigma _{x}^{2} & {{\sigma }_{xy}}  \\
   {{\sigma }_{yx}} & \sigma _{y}^{2}  \\
\end{matrix} \right]
\label{eq3}
\end{equation}

The LED marker has a spherical design, producing a typical circular shape on the projection plane of the camera, so ${{\sigma }_{xy}}={{\sigma }_{yx}}=0$. Accumulate $m$ events and calculate the centroid of these events. High-frequency blinking markers trigger events intensively through active changes in light intensity. When extracting the marker center point, the positional information from a single event is insufficient to reflect changes in the marker position. Therefore, a fixed number of events is accumulated, and the positions of the event clusters are extracted as the marker center point. At this stage, the timestamp of an individual event does not accurately represent the time at which the center point is extracted. If $n$ events are accumulated to extract the marker center point, the time $t_{c}$ can be represented as Eq. (\ref{eq6}):
\begin{equation}
t_{c}=\frac{1}{n} \sum_{i=1}^{n} t_{i},
\label{eq6}
\end{equation}
where $t_{i}$ represents the timestamp of each accumulated event, and $t_{c}$ is the time to extract the marker center points.

Event stream data differs from traditional image frames because it is asynchronous and sparse. A single event provides minimal information, and extracting the center of the marker requires accumulating and processing multiple events. At specific pixel locations, a change in brightness triggers an event in two event cameras. The timing of these output events does not match the exact time of real brightness changes in the physical world. There is a delay in the arrival of changes in light intensity at the pixel position. Additionally, spatial aliasing can also affect the timing of events, making it impossible to match multiple event camera pixels based on the exact timing of the output events.

\begin{figure}[htbp]
\centering\includegraphics[width=8cm]{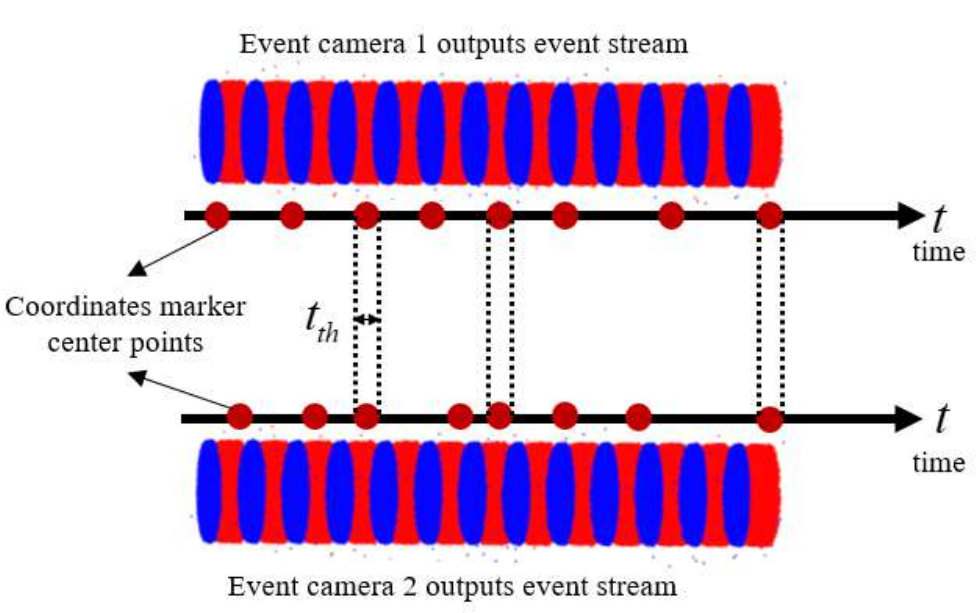}
\caption{Corresponding points extraction from event streams generated by various event cameras.}
\label{fig2}
\end{figure}

This paper explores the use of temporal correlation to achieve high-precision matching of multi-event cameras and extract corresponding points, as illustrated in Fig.~\ref{fig2}. Event camera 1 extracts the marker center $(x_{i},y_{i})$ at time $t_{1}$, and camera 2 extracts $(x_{j},y_{j})$ at $t_{2}$. Assuming the time threshold is $t_{th}$:
\begin{equation}
\left|t_{1}-t_{2}\right| \leq t_{t h},
\label{eq7}
\end{equation}
then it is considered that $(x_{i},y_{i})$ and $(x_{j},y_{j})$ are the corresponding points as the markers. The time threshold $t_{th}$ serves to establish correspondences between event streams. Its value depends on factors such as the marker's velocity and the synchronization accuracy of the event cameras. Usually, a very small time threshold, on the order of $10^{-8}$ seconds, is required to ensure precise matching of corresponding points. Due to differing requirements between calibration and measurement tasks, the marker extraction process differs during deformation measurement, as detailed in Section \ref{sec2.3}. 

\subsection{Fast Self Calibration of Multi-Event Camera System}\label{sec2.2}
Assume there are $m$ event cameras and $n$ object points $\mathbf{x}_{j}=\left[X_{j}, Y_{j}, Z_{j}, 1\right]^{^\top}, j=1, \ldots, n$. The correspondence between the pixel positions of object points and their corresponding events ${{\mathbf{x}}_{j}}={{[{{u}_{j}},{{v}_{j}},1]}^{^\top}},j=1,...,n$ can be obtained from the pinhole imaging model. Event camera calibration involves calculating the scale factor $s$ and specific parameters of the camera projection matrix $\mathbf{M}$. Express the relationship between all cameras and points using Eq. (\ref{eq8})~\cite{bib29}:
\begin{eqnarray}
{\mathbf{W}_{s}}&=&\left[ \begin{matrix}
   s_{1}^{1}\left[ \begin{matrix}
   u_{1}^{1}  \\
   v_{1}^{1}  \\
   1  \\
\end{matrix} \right] & \cdots  & s_{n}^{1}\left[ \begin{matrix}
   u_{n}^{1}  \\
   v_{n}^{1}  \\
   1  \\
\end{matrix} \right]  \\
   \vdots  & \vdots  & \vdots   \\
   s_{1}^{m}\left[ \begin{matrix}
   u_{1}^{m}  \\
   v_{1}^{m}  \\
   1  \\
\end{matrix} \right]\cdots  & {} & s_{n}^{m}\left[ \begin{matrix}
   u_{n}^{m}  \\
   v_{n}^{m}  \\
   1  \\
\end{matrix} \right]  \\
\end{matrix} \right]   \nonumber\\
&=&{{\left[ \begin{matrix}
   {\mathbf{M}^{\text{l}}}  \\
   \vdots   \\
   {\mathbf{M}^{m}}  \\
\end{matrix} \right]}_{3m\times 4}}{{\left[ {{\mathbf{X}}_{1}}\cdots {{\mathbf{X}}_{n}} \right]}_{4\times n}}=\mathbf{MX},
\label{eq8}
\end{eqnarray}
where $\mathbf{W}_{s}$ are called measurement matrix, $\mathbf{M}={{[{\mathbf{M}^{1}},...,{\mathbf{M}^{m}}]}^{\top}}$ is camera matrix, and $X=\left[ {{\mathbf{X}}_{1}}\cdots {{\mathbf{X}}_{n}} \right]$ is structure matrix. If there are enough known coordinates ${{[u,v,1]}^{\top}}$ of the pixels and scale factors $s$, the matrix $\mathbf{W}_{s}$ can be decomposed into $\mathbf{M}$ and $\mathbf{X}$~\cite{bib29}.

Assuming all initial scale factor values are 1, obtain the initial value of $\mathbf{W}_{s}$. Perform SVD decomposition on $\mathbf{W}_{s}$:
\begin{equation}
{{\mathbf{W}}_{s}}=\mathbf{UD{{V}}^{\top}}
\label{eq9}
\end{equation}

For an ideal situation, the rank of the measurement matrix $\mathbf{W}_{s}$ is 4. However, due to data errors, the rank of $\mathbf{W}_{s}$ is not 4. Therefore, rank 4 correction is applied to the decomposition result of $\mathbf{W}_{s}$ as the best approximation of the measurement matrix:
\begin{equation}
\widehat{{\mathbf{{W}}_{s}}}=\mathbf{U\widehat{D}{{V}}^{\top}},
\label{eq10}
\end{equation}
where $\mathbf{\widehat{D}}$ is the matrix obtained by taking zeros for all elements of $\mathbf{D}$ except for the first four diagonal elements. Therefore, the photographic reconstruction of the camera matrix is:
\begin{equation}
\tilde{\mathbf{M}} = [\tilde{\mathbf{M}}^1, \ldots, \tilde{\mathbf{M}}^m]^{\mathrm{T}} = \mathbf{U}\tilde{\mathbf{D}}^{\top}
\label{eq11}
\end{equation}

The projective reconstruction of the structural matrix is:
\begin{equation}
\widetilde{\mathbf{X}}=[{{\widetilde{\mathbf{X}}}_{1}}\cdots {{\widetilde{\mathbf{X}}}_{n}}]={{\mathbf{V}}^{\top}}
\label{eq12}
\end{equation}

Reproject the results of the photographic reconstruction of the structural matrix and the camera matrix to obtain new estimates for each scale factor. Substitute the new scale factor to calculate the measurement matrix, repeat the above steps until convergence, and obtain the projection reconstruction result of the spatial target point.

If the event camera collects data from directions $j_{0}$ and $j_{1}$ respectively, there exists the Kruppa equation as follows:
\begin{equation}
\mathbf{FC{{F}}^{\top}}=s{{\left[ {{e}_{j1}} \right]}_{\times }}\mathbf{C}\left[ {{e}_{j1}} \right]_{\times }^{\top},
\label{eq13}
\end{equation}
where $s$ is the scaling factor, $\mathbf{F}$ is the basic matrix of two directional cameras, and $e_{j1}$ is the homogeneous coordinate of the image poles of the event camera collecting event data in the $j_{1}$ direction. $\mathbf{C}=\mathbf{K{{K}}^{\top}}$, $\mathbf{K}$ is the parameter matrix inside the camera. ${{\left[ {{e}_{j1}} \right]}_{\times }}$ is an antisymmetric matrix defined by the vector ${{e}_{j1}}=[{e}_{j1,0} {e}_{j1,1} {e}_{j1,2}]^{\top}$:
\begin{equation}
{{\left[ {{\mathbf{e}}_{j1}} \right]}_{\times }}=\left[ \begin{matrix}
   0 & -{{e}_{j1,2}} & {{e}_{j1,1}}  \\
   {{e}_{j1,2}} & 0 & -{{e}_{j1,0}}  \\
   -{{e}_{j1,1}} & {{e}_{j1,0}} & 0  \\
\end{matrix} \right]
\label{eq14}
\end{equation}

Keep the camera's internal parameters unchanged. Suppose the initial value of the camera's principal point is taken at the center of the image plane, and the initial values of the equivalent focal length in the horizontal and vertical directions are the same, denoted as $f$. In that case, the only internal parameter to be solved is the equivalent focal length $f$. At this point, the equivalent focal length $f$ can be easily solved from the Kruppa equation.
\begin{eqnarray}
\mathbf{FC{{F}}^{\top}}=s\left[ \begin{matrix}
   0 & -{{e}_{j1,2}} & {{e}_{j1,1}}  \\
   {{e}_{j1,2}} & 0 & -{{e}_{j1,0}}  \\
   -{{e}_{j1,1}} & {{e}_{j1,0}} & 0  \\
\end{matrix} \right] \nonumber \\ \left[ \begin{matrix}
   {{f}^{2}}+C_{x}^{2} & {{C}_{x}}{{C}_{y}} & {{C}_{x}}  \\
   {{C}_{x}}{{C}_{y}} & {{f}^{2}}+C_{y}^{2} & {{C}_{y}}  \\
   {{C}_{x}} & {{C}_{y}} & 1  \\
\end{matrix} \right] \nonumber \\
{{\left[ \begin{matrix}
   0 & -{{e}_{j1,2}} & {{e}_{j1,1}}  \\
   {{e}_{j1,2}} & 0 & -{{e}_{j1,0}}  \\
   -{{e}_{j1,1}} & {{e}_{j1,0}} & 0  \\
\end{matrix} \right]}^{\top}}
\label{eq15}
\end{eqnarray}

Expanding the equation to obtain Eq. (\ref{eq16}).                                          
\begin{equation}
\mathbf{FC{{F}}^{T}}=s\mathbf{D}=s\left[ \begin{array}{*{35}{l}}
   {{D}_{11}} & {{D}_{12}} & {{D}_{13}}  \\
   {{D}_{21}} & {{D}_{22}} & {{D}_{23}}  \\
   {{D}_{31}} & {{D}_{32}} & {{D}_{33}}  \\
\end{array} \right],
\label{eq16}
\end{equation}

Dividing the first two columns by the third column can eliminate the scaling factor $s$, thereby obtaining a system of linear equations about ${{f}^{2}}$ and obtaining the equivalent focal length $f$.

Due to the possibility that the projection reconstruction results may differ from the real camera parameters and target structure by a 3D projection transformation, for any 4 × 4 nonsingular matrix $\mathbf{H}$, the projection reconstruction results $\widetilde{\mathbf{M}}$ and $\widetilde{\mathbf{X}}$, as well as the Euclidean reconstruction results $\widehat{\mathbf{M}}$ and $\widehat{\mathbf{X}}$, satisfy the following conditions:
\begin{equation}
{{\mathbf{W}}_{s}}=\widetilde{\mathbf{M}}\widetilde{\mathbf{X}}=\mathbf{\widetilde{M}H{{H}}^{-1}}\mathbf{\widetilde{X}}=\mathbf{\widehat{M}\widehat{X}},
\label{eq18}
\end{equation}

To further obtain the Euclidean reconstruction result that describes the true target structure from the projection reconstruction result, $\mathbf{H}$ should satisfy $\mathbf{\widehat{M}=MH}$ and $\mathbf{\widehat{X}={{H}^{-1}}X}$.

According to the central perspective projection relationship, the Euclidean reconstruction form of the camera matrix at the $j$ orientation should be:
\begin{equation}
{{\widehat{\mathbf{M}}}_{j}}={{\widetilde{\mathbf{M}}}_{j}}\mathbf{H}={{\mathbf{K}}_{j}}\left[ \begin{array}{*{35}{l}}
   {{\mathbf{R}}_{j}} & {{\mathbf{T}}_{j}}  \\
\end{array} \right],
\label{eq19}
\end{equation}
where ${{\mathbf{K}}_{j}}$ is the internal parameter matrix of the event camera at the $j$ orientation, and $\mathbf{R}_{j}$, $\mathbf{T}_{j}$ are the rotation matrix and translation vector of the event camera at the $j$ orientation

Rewrite the 4 $\times$ 4 $\mathbf{H}$ as
\begin{equation}
\mathbf{H}=\left[ \begin{array}{*{35}{l}}
   {{\mathbf{H}}_{11}} & {{\mathbf{h}}_{12}}  \\
\end{array} \right],
\label{eq20}
\end{equation}
where ${{\mathbf{H}}_{11}}$ is a 4 $\times$ 3 order matrix composed of the first three columns of $\mathbf{H}$ elements, and $\mathbf{h}_{12}$ is the last column of $\mathbf{H}$, thus obtaining:
\begin{equation}
{{\widetilde{\mathbf{M}}}_{j}}{{\mathbf{H}}_{11}}={{\mathbf{K}}_{j}}{{\mathbf{R}}_{j}}
\label{eq21}
\end{equation}

The transpose on both sides of the equal sign is:
\begin{equation}
\mathbf{H}_{11}^{\top}\widetilde{\mathbf{M}}_{j}^{\top}=\mathbf{R}_{j}^{\top}\mathbf{K}_{j}^{\top}
\label{eq22}
\end{equation}

Since the rotation matrix ${{\mathbf{R}}_{j}}$ is a unit orthogonal matrix, there are:
\begin{equation}
{{\widetilde{\mathbf{M}}}_{j}}{{\mathbf{H}}_{11}}\mathbf{H}_{11}^{\top}\widetilde{\mathbf{M}}_{j}^{\top}={{\mathbf{K}}_{j}}\mathbf{RR}_{j}^{\top}\mathbf{K}_{j}^{\top}={{\mathbf{K}}_{j}}\mathbf{K}_{j}^{\top}
\label{eq23}
\end{equation}

If $\mathbf{G}={{\mathbf{H}}_{11}}\mathbf{H}_{11}^{T}$, then $\mathbf{G}$ is a 4 $\times$ 4 symmetric matrix, and $\mathbf{H}$ is a 4 $\times$ 3 matrix. Therefore, by performing SVD decomposition on $\mathbf{G}$:
\begin{eqnarray}
\mathbf{G} &=& \mathbf{A}\left[ \begin{array}{*{35}{l}}
   {{\sigma }_{0}} & {} & {} & {}  \\
   {} & {{\sigma }_{1}} & {} & {}  \\
   {} & {} & {{\sigma }_{2}} & {}  \\
   {} & {} & {} & 0  \\
\end{array} \right]{{A}^{\top}} \nonumber \\
& =& \mathbf{A}\left[ \begin{array}{*{35}{l}}
   \sqrt{{{\sigma }_{0}}} & {} & {} & {}  \\
   {} & \sqrt{{{\sigma }_{1}}} & {} & {}  \\
   {} & {} & \sqrt{{{\sigma }_{2}}} & {}  \\
   {} & {} & {} & 0  \\
\end{array} \right] \nonumber \\
& &{{\left[ \begin{array}{*{35}{l}}
   \sqrt{{{\sigma }_{0}}} & {} & {} & {}  \\
   {} & \sqrt{{{\sigma }_{1}}} & {} & {}  \\
   {} & {} & \sqrt{{{\sigma }_{2}}} & {}  \\
   {} & {} & {} & 0  \\
\end{array} \right]}^{\top}}{{\mathbf{A}}^{\top}}
\label{eq24}
\end{eqnarray}

If $\mathbf{A}_{11}$ is a $4\times 3$ matrix composed of the first three columns of elements $\mathbf{A}$, we can obtain:
\begin{equation}
\mathbf{H}_{11} = \mathbf{A}_{11}
\begin{bmatrix}
\sqrt{\sigma_0} & & \\
& \sqrt{\sigma_1} & \\
& & \sqrt{\sigma_2}
\end{bmatrix}
\label{eq25}
\end{equation}

If any point in $\mathbf{X}$ is taken as the origin of Euclidean coordinates, then:
\begin{equation}
{{\left[ \begin{array}{*{35}{l}}
   0 & 0 & 0 & 1  \\
\end{array} \right]}^{\top}}={{\mathbf{H}}^{-1}}{{\widetilde{\mathbf{X}}}_{i0}}
\label{eq26}
\end{equation}

So the last column $\mathbf{h}_{12}$ is:
\begin{equation}
{{\mathbf{h}}_{12}}={{\widetilde{\mathbf{X}}}_{i0}}
\label{eq27}
\end{equation}

Thus, the $\mathbf{H}$ matrix was obtained. Substitute the $\mathbf{H}$ matrix and projection reconstruction results $\widetilde{\mathbf{M}}$ and $\widetilde{\mathbf{X}}$ into Eq. (\ref{eq18}) to obtain the 3D coordinates of the event camera parameters and target feature points, as well as the Euclidean reconstruction results $\widehat{\mathbf{M}}$ and $\widehat{\mathbf{X}}$. The self-calibration result lacks scale information.

First, compute the projection scale factor. Taking event camera $c$ as the center, let the scale factor for all known corresponding point pairs $p$ projected onto this event camera be $s_p^c = 1$. For another event camera $i$, if it shares sufficient corresponding point pairs with event camera $c$ to compute a unique fundamental matrix, the scale factor $s_p^i$ for event camera $i$ can be calculated using Eq. (\ref{eq28})~\cite{bib29}:
\begin{equation}
s_{p}^{i} = \frac{\left( \mathbf{e}^{ic} \times \mathbf{u}_{p}^{i} \right) \times \left( \mathbf{F}^{ic} \mathbf{u}_{p}^{c} \right)}{\left\| \mathbf{e}^{ic} \times \mathbf{u}_{p}^{i} \right\|^{2}} s_{p}^{c},
\label{eq28}
\end{equation}
where $\mathbf{F}^{ic}$ denotes the fundamental matrix, $\mathbf{e}^{ic}$ represents the epipole, and $\mathbf{u}_p^c$ and $\mathbf{u}_p^i$ indicate the projected pixel coordinates on event camera $c$ and event camera $i$, respectively.

Next, the projection structure is optimized using Bundle Adjustment. From Eq. (\ref{eq8}), the projection relationship of a 3D point onto the camera coordinate system is expressed in matrix form as:
\begin{equation}
s_j \mathbf{u}_j = \mathbf{K} \exp(\xi^\wedge) \mathbf{X}_j,
\label{eq29}
\end{equation}
where $\mathbf{u}_j$ represents the projection of the 3D point on the camera, and $\exp(\xi^\wedge)$ denotes the camera's extrinsic parameters. Due to unknown camera poses and noise in observed points, this equation contains errors. Summing these errors, a least squares problem is formulated to find the optimal camera pose that minimizes the total error, as shown in Eq. (\ref{eq30}):
\begin{equation}
\xi^* = \arg \min_{\xi} \frac{1}{2} \sum_{j=1}^n \left\| \mathbf{u}_j - \frac{1}{s_j} \mathbf{K} \exp(\xi^\wedge) \mathbf{X}_j \right\|_2^2
\label{eq30}
\end{equation}

According to subsection \ref{sec2.2}, the measurement matrix is decomposed to obtain projective reconstruction results for the camera matrix and structure matrix. Euclidean reconstruction is then applied to recover the true camera parameters.  

Next, outlier projection points are removed. Despite the robustness of the proposed cooperative marker extraction method, outliers may persist due to factors such as irrelevant background clutter, specular reflections causing high-frequency marker halo artifacts, and sensor-induced noise (e.g., background or thermal noise). These outliers degrade calibration accuracy or may even cause failure. The proposed method eliminates outliers by computing epipolar geometric errors and reprojection errors of dynamic markers.  

For the extracted point sequences from two cameras, the epipolar lines are robustly estimated using the random sample consensus (RANSAC) seven-point algorithm~\cite{bib30}. Let $\mathbf{K}_1$ and $\mathbf{K}_2$ denote the intrinsic parameter matrices of camera 1 and camera 2, respectively, and $\mathbf{R}, \mathbf{t}$ represent the geometric transformation parameters (rotation and translation) from camera 1 to camera 2. The fundamental matrix $\mathbf{F}$ between the cameras is computed as: 
\begin{equation}
\mathbf{F} = \mathbf{K}_2^{-\top} [\mathbf{t}]_\times \mathbf{R} \mathbf{K}_1^{-1}
\label{eq31}
\end{equation}

The fundamental matrix $\mathbf{F}$ contains 7 degrees of freedom, requiring at least 7 matched feature pairs for estimation. Let $\mathbf{u}_1$ and $\mathbf{u}_2$ be the homogeneous coordinates of a 3D point projected onto event camera 1 and 2, respectively. The epipolar line $\mathbf{l}_2$ on camera 2 is derived from $\mathbf{F}\mathbf{u}_1 = [A, B, C]^T$, yielding the equation:  
\begin{equation}
A u + B v + C = 0 
\label{eq32}
\end{equation}

A distance threshold $d_h$ is defined for outlier rejection. Points violating the following condition are discarded:  
\begin{equation}
\frac{|\mathbf{u}_2^T \mathbf{F} \mathbf{u}_1|}{\sqrt{A^2 + B^2}} > d_h,
\label{eq33}
\end{equation}
where points $\mathbf{u}_1$ and $\mathbf{u}_2$ are identified as outliers and removed.  

Let $\xi_{\text{th}}$ denote the reprojection error threshold for projected points on each camera. If the optimized reprojection error satisfies:
\begin{equation}
\xi^* > \xi_{\text{th}},
\label{eq34}
\end{equation}
then the corresponding projected point is classified as an outlier and discarded.

Finally, the distortion parameters are estimated. First, the dynamic markers are reconstructed using the linear camera parameters obtained from previous steps. These 3D-2D correspondences are fed into a standard method to compute the camera distortion parameters. The linear self-calibration is repeated using the corrected point coordinates, iterating until the desired accuracy is achieved. The summary of the calibration algorithm steps is shown in Table~\ref{Algorithm}.

\begin{table*}[htbp]
\caption{Calibration algorithm.}\label{Algorithm}
\centering
\resizebox{\linewidth}{!}{
\begin{tabular}{c p{0.8\linewidth}}
\toprule
\textbf{Input} & Corresponding points for each camera $\mathbf{P}$, Iteration count $i$, Initial value of camera projection matrix $\mathbf{M}$ \\ 
\textbf{Output} & Intrinsic parameters $\mathbf{K}$, extrinsic parameters $\mathbf{R,T}$ \\
\midrule
1 & Calculate scale factor \\
2 & Optimize the projection structure using the bundle adjustment method \\
3 & Decompose the measurement matrix to obtain the projection reconstruction of the camera matrix and structure matrix \\
4 & Perform Euclidean reconstruction to obtain camera parameters \\
5 & By calculating the reprojection errors of polar geometry and dynamic landmark points, outlier projection points are eliminated \\
6 & Solving distortion parameters \\
7 & Repeat steps 1-5 to obtain the optimal camera parameters. When the average reprojection error of all points in each camera is less than the given threshold or reaches the maximum number of iterations, the iteration is terminated \\
\bottomrule
\end{tabular}}
\end{table*}

The complexity of the nonlinear model gradually increases during iteration. However, the distortion parameter estimation may fail if the markers are unevenly distributed in the camera’s field of view or excessive outliers persist in marker extraction. In that case, reducing the nonlinear model complexity and restricting the number of free parameters is advisable. For instance, when solving for the principal point, the estimation becomes unstable if markers are clustered on one side of the image. In such cases, halting distortion parameter computation and fixing the principal point at the image center is preferable. While this may result in higher final reprojection errors, forcing erroneous nonlinear parameter estimates could compromise geometric consistency.  

The self-calibration process iteratively refines camera parameters until the mean reprojection error across all points falls below a predefined threshold or the maximum iteration count is reached, thereby completing the self-calibration of the event camera.  

\subsection{3D Deformation Measurement Utilizing Multi-event Cameras}\label{sec2.3}
To enable high-speed deformation measurements of structures under harsh lighting conditions, robust marker extraction must first be achieved. This requires not only suppressing interference from ambient background noise to ensure marker extraction accuracy, but also applying specialized processing—as outlined in Section \ref{sec2.1}—to the event stream generated by rapidly moving markers.  As is seen in Eq. (\ref{eq2}), the computation of the marker center coordinates requires simultaneous processing of $n$ events. To ensure accurate extraction of rapidly moving markers, the value of $n$ is governed by Eq. (\ref{eq:eqnew}):
\begin{equation}
n = f(\nu,v)
\label{eq:eqnew}
\end{equation}
where $\nu$ is the blinking frequency of the LED marker, and $v$ represents its velocity. The value of $n$ increases with the LED marker's blinking frequency and decreases with its displacement rate. Consequently, calibration uses a lower blinking frequency to prioritize speed. In deformation measurement, the movement speed of markers is generally higher, and in order to distinguish them from events caused by structure movement, the blinking frequency needs to be higher. The specific value of n is dynamically determined based on these experimental conditions.
\begin{figure}[ht]
    \centering
    \includegraphics[width=9cm]{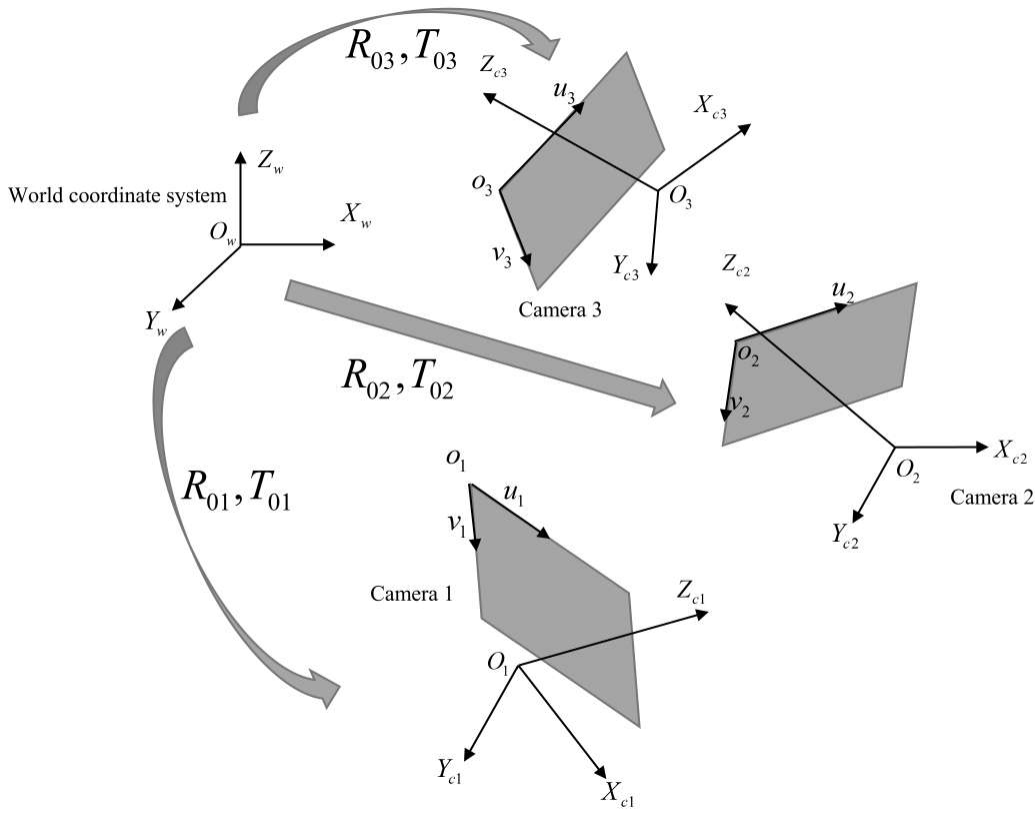}
    \caption{Reference coordinate system transformation.}
    \label{fig3}
\end{figure}

The fast self-calibration method in Section \ref{sec2.2} enables rapid estimation of camera pose relative to the world coordinate system. For deformation measurement applications, careful selection of the world coordinate system is essential, necessitating an initial transformation of the reference coordinate system. As is seen in Fig.~\ref{fig3}, the world coordinate system is defined as $O_w-X_wY_wZ_w$. Meanwhile, the camera coordinate system of camera 1 is $O_1-X_{c1}Y_{c1}Z_{c1}$, and the camera coordinate system of camera 1 and camera 2 are $O_2-X_{c2}Y_{c2}Z_{c2}$ and $O_3-X_{c3}Y_{c3}Z_{c3}$, respectively. Among them, the rotation matrix and the direction vectors from the world coordinate system to $O_1-X_{c1}Y_{c1}Z_{c1}$ are $\mathbf{R}_{01}$ and $\mathbf{T}_{01}$; similarly, the direction vectors to $O_2-X_{c2}Y_{c2}Z_{c2}$ and $O_3-X_{c3}Y_{c3}Z_{c3}$ are $\mathbf{R}_{02}$, $\mathbf{T}_{02}$ and $\mathbf{R}_{03}$, $\mathbf{T}_{03}$, respectively.

To simplify the calculation process, we choose the coordinate system of camera 1 as the reference coordinate system. At this point, we need to separately calculate the rotation matrix and translation vector of camera 2's coordinate system and camera 3's coordinate system relative to camera 1's coordinate system, as is seen in Eq. (\ref{eq35}) and Eq. (\ref{eq36}):
\begin{equation}
\mathbf{R}_{12}=\mathbf{R}_{02} \cdot \mathbf{R}_{01}^{\top}, \mathbf{R}_{13}=\mathbf{R}_{03} \cdot \mathbf{R}_{01}^{\top},
\label{eq35}
\end{equation}
\begin{equation}
\mathbf{T}_{12}=\mathbf{T}_{02} - \mathbf{R}_{12} \cdot \mathbf{T}_{01}, \mathbf{T}_{13}=\mathbf{T}_{03} - \mathbf{R}_{13} \cdot \mathbf{T}_{01},
\label{eq36}
\end{equation}
where $\mathbf{R}_{12}$ represents the direct rotation from camera 1 to camera 2, and $\mathbf{T}_{12}$ is the offset of camera 2's origin relative to camera 1. Similarly, $\mathbf{R}_{13}$ represents the direct rotation from camera 1 to camera 3, and $\mathbf{T}_{13}$ is the offset of camera 3's origin relative to camera 1.

The planar intersection algorithm leverages knowledge of geometry to determine the geometric relationship between the camera, the imaging plane, and the target. The first camera coordinate system is defined as $ O_{1}-X_{c 1} Y_{c 1} Z_{c 1}$, and the corresponding image frame as $ o_{1}-u_{1}v_{1}$. Similarly, the second camera coordinate system and the corresponding image frame coordinate system are defined separately as  $ O_{2}-X_{c 2} Y_{c 2} Z_{c 2}$ and $ o_{2}-u_{2}v_{2}$. The third camera coordinate system and the corresponding image frame coordinate system are defined separately as  $ O_{3}-X_{c 3} Y_{c 3} Z_{c 3}$ and $ o_{3}-u_{3}v_{3}$. In the absence of an inclinometer to determine the spatial attitude of the camera, it is assumed that the left camera coordinate system is the reference coordinate system. Thus, coordinates of the points on the left and right camera imaging planes can be expressed using Eq. (\ref{eq:eq1}):
\begin{equation}
 Z_{c1}\left[\begin{array}{c}u_{1} \\v_{1} \\1\end{array}\right]=\mathbf{M}\left[\begin{array}{c}X \\Y \\Z \\1\end{array}\right], \ \
  Z_{c 2}\left[\begin{array}{c}u_{2} \\v_{2} \\1\end{array}\right]=\mathbf{N}\left[\begin{array}{c}X \\Y \\Z \\1\end{array}\right],
    Z_{c 3}\left[\begin{array}{c}u_{3} \\v_{3} \\1\end{array}\right]=\mathbf{P}\left[\begin{array}{c}X \\Y \\Z \\1\end{array}\right],
\label{eq:eq1}
\end{equation}
where $Z_{c 1}$, $Z_{c 2}$ and $Z_{c3}$ represent the projections of the distance from the object points to the optical center in the direction of the optical axis. $ (u_{1},v_{1})$, $(u_{2},v_{2})$and $(u_{3},v_{3})$ are the ideal image coordinates of the image points on the each camera image frames, respectively. $(X,Y,Z)$ represents the world coordinates of the object point. $\mathbf{M}$ is the projection matrix of the first camera, while $\mathbf{N}$ is the projection matrix of the second camera. $\mathbf{P}$ represents the projection matrix of the third camera. The matrix $\mathbf{M}$, $\mathbf{N}$ and $\mathbf{P}$ can be obtained from the Eq. (\ref{eq:eq2}):
 \begin{equation}
    \mathbf{M}={\mathbf{A}_{1}}\left[ \begin{matrix}
   \mathbf{I} & \mathbf{0}  \\
   {\mathbf{0}^{\top}} & 1  \\
\end{matrix} \right], \ \
\mathbf{N}={\mathbf{A}_{2}}\left[ \begin{matrix}
   \mathbf{R}_{12} & \mathbf{T}_{12}  \\
   {\mathbf{0}^{\top}} & 1  \\
\end{matrix} \right],
\mathbf{P}={\mathbf{A}_{3}}\left[ \begin{matrix}
   \mathbf{R}_{13} & \mathbf{T}_{13}  \\
   {\mathbf{0}^{\top}} & 1  \\
\end{matrix} \right],
\label{eq:eq2}
\end{equation}
where $\mathbf{A_{1}}$, $\mathbf{A_{2}}$ and $\mathbf{A_{3}}$ are the intrinsic parameter matrices of the each camera, respectively. $\mathbf{I}$ is a 3×3 identity  matrix. $\mathbf{R}_{12}$, $\mathbf{T}_{12}$ and $\mathbf{R}_{13}$, $\mathbf{T}_{13}$ can be obtained by Eq. (\ref{eq35}) and Eq. (\ref{eq36}). It is possible to measure the 3D position $(X,Y,Z)$ of markers by combining Eq. (\ref{eq2}), Eq. (\ref{eq:eqnew}) and Eq. (\ref{eq:eq1}):
\begin{equation}
\frac{ Z_{c1} }{n} \mathbf{M}^{-1} \left[\begin{array}{c}\sum\limits_{i=1}^{n}{{{u}_{1i}}} \\ \sum\limits_{i=1}^{n}{{{v}_{1i}}} \\1\end{array}\right]= \ \
\frac{ Z_{c2} }{n} \mathbf{N}^{-1} \left[\begin{array}{c}\sum\limits_{i=1}^{n}{{{u}_{2i}}} \\ \sum\limits_{i=1}^{n}{{{v}_{2i}}} \\1\end{array}\right]= \ \
\frac{ Z_{c3} }{n} \mathbf{P}^{-1} \left[\begin{array}{c}\sum\limits_{i=1}^{n}{{{u}_{3i}}} \\ \sum\limits_{i=1}^{n}{{{v}_{3i}}} \\1\end{array}\right]= \ \
\left[\begin{array}{c}X \\Y \\Z \\1\end{array}\right],
\label{eq:eqnew2}
\end{equation}

\section{Experiments}\label{sec3}
\begin{figure}[ht]
\centering
\subfigure[Blinking LED marker.]{ 
\begin{minipage}{0.5\linewidth}
\centering 
\includegraphics[height = 4.5cm]{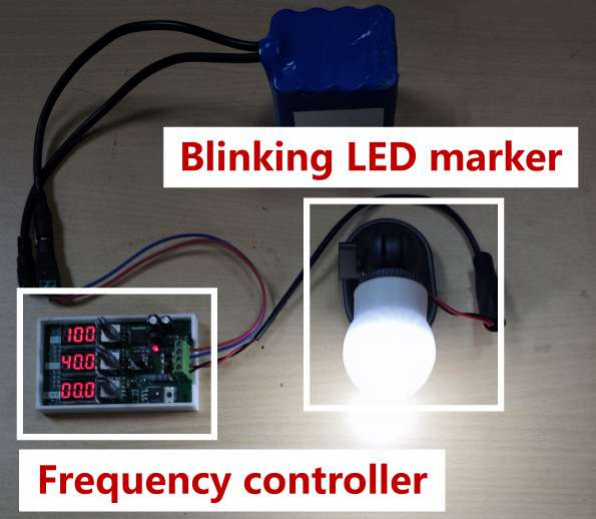}  
\end{minipage}
\label{fig4a}
}\subfigure[Configuration of the multi-event camera array.]{ 
\begin{minipage}{0.5\linewidth}
\centering  
\includegraphics[height = 4.5cm]{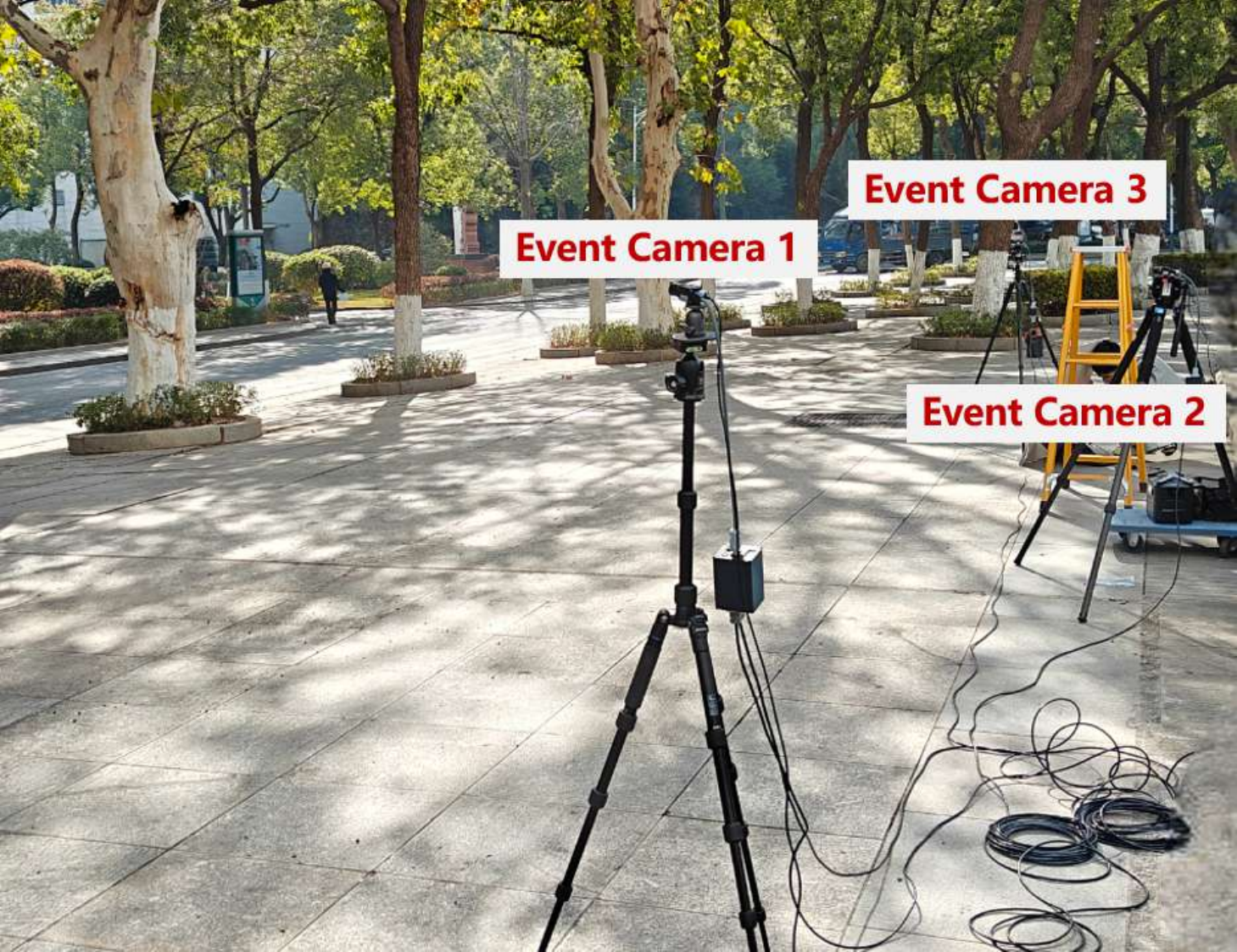}
\end{minipage}
\label{fig4b}
}
\caption{Experimental configuration.} 
\label{fig4}   
\end{figure}

To verify the proposed method, experiments were conducted as shown in Fig.~\ref{fig4}. Fig.~\ref{fig4a} displays the blinking LED marker setup, used for calibration and deformation measurement. The spherical marker has a 5cm outer diameter. Its frequency controller sets the LED between 1-100 kHz and adjusts the duty cycle. Empirical testing determined a 250 Hz frequency and 40$\%$ duty cycle as optimal for camera event stream quality.

As is seen in Fig.~\ref{fig4b}, three event cameras formed an array in the experiment. The model of the camera is Prophete EVK4 (resolution 1280 × 720 pixels), and the camera is equipped with a lens with a focal length of 12$mm$. Three cameras can collect data synchronously. The event cameras were mounted on tripods. The baseline distance between camera 1 and camera 2 is 4.64m, and the baseline distance between camera 2 and camera 3 is 4.54m. 

\subsection{Fast Self-calibration of the Multi-event Camera Array}\label{sec3.1}
Prior to structural deformation measurements, calibrate the event camera array depicted in Fig.~\ref{fig4b}. Use the dynamic high-frequency blinking LED marker to move in the public field of view of the event camera system and create a dynamic calibration field. Then, use a synchronized multi-event camera array to collect spatiotemporal information of the marker. Once the markers are extracted via the proposed method, proceed to calibrate the camera array.

In the experiment, the initial camera intrinsics $({{f}_{x}},{{f}_{y}})$ is set as $(1600,1600)$, and $({{C}_{x}},{{C}_{y}})$ is set as $(639.5,359.5)$. The optical center can be assumed to be located at the center of the image, and the tilt factor of the camera is 0. Set the initial distortion parameter to 0. Set the maximum number of iterations to 20 or stop iterating when the average of all reprojections for each camera is less than 0.3 pixels. The results of calibrating the internal parameters of the event cameras using the proposed method are shown in Table~\ref{tab1}.

\begin{table*}[htbp]
\caption{Calibration results of each camera intrinsics.}\label{tab1}
    \centering
    \resizebox{\linewidth}{!}{
    \begin{tabular}{c c c c}
\toprule
Extrinsics & Event camera 1 & Event camera 2 & Event camera 3 \\ 
\midrule
$(f_x, f_y)$ & (1778.5077, 1772.3397) & (1756.2173, 1757.6988) & (1816.1340, 1809.6153) \\ 
$(C_x, C_y)$ & (639.5, 359.5) & (639.5, 359.5) & (639.5, 359.5) \\ 
$(k_1, k_2)$ & (-0.05359, 0.33899) & (-0.09119, 0.38824) & (-0.11437, 0.91938) \\ 
$(p_1, p_2)$ & (-0.00157, -0.00479) & (0.00149, -0.00318) & (-0.00144, 0.00497) \\ 
\botrule  
\end{tabular}}
\end{table*}

The definition of the world coordinate system is random. However, it is hoped that the defined world coordinate system can simplify the calculation process and lend practical significance to the calculated parameters. We set the coordinate system of camera 2 as the reference coordinate system. Then, the extrinsics of camera 1 and camera 3 were calibrated separately. The results are shown in Table~\ref{tab2}. The translation vector here represents the relative positional relationship between the cameras and does not have any actual physical scale meaning.
\begin{table*}[htbp]
\caption{Calibration results of each camera extrinsics.}\label{tab2}
    \centering
    \resizebox{\linewidth}{!}{
    \begin{tabular}{c c c c}
\toprule
Extrinsics & Event camera 2 & Event camera 1 & Event camera 3 \\ 
\midrule
$R$ & $\begin{bmatrix} 1 & 0 & 0 \\ 0 & 1 & 0 \\ 0 & 0 & 1 \end{bmatrix}$ & $\begin{bmatrix} 0.99267 & 0.00302 & 0.06502 \\ 0.01514 & 0.86947 & -0.50846 \\ -0.09079 & 0.51537 & 0.85804 \end{bmatrix}$ & $\begin{bmatrix} 0.97156 & 0.09647 & 0.02984 \\ 0.00481 & 0.29765 & -0.93251 \\ -0.11474 & 0.95562 & 0.30323 \end{bmatrix}$ \\ 
$T$ & $\begin{bmatrix} 0 \\ 0 \\ 0 \end{bmatrix}$ & $\begin{bmatrix} 0.2224 \\ -0.8944 \\ -0.3881 \end{bmatrix}$ & $\begin{bmatrix} 0.1517 \\ -0.5671 \\ -0.8096 \end{bmatrix}$ \\ 
\botrule  
\end{tabular}}
\end{table*}

\begin{figure}[htbp]
    \centering
    \includegraphics[width=0.4\linewidth]{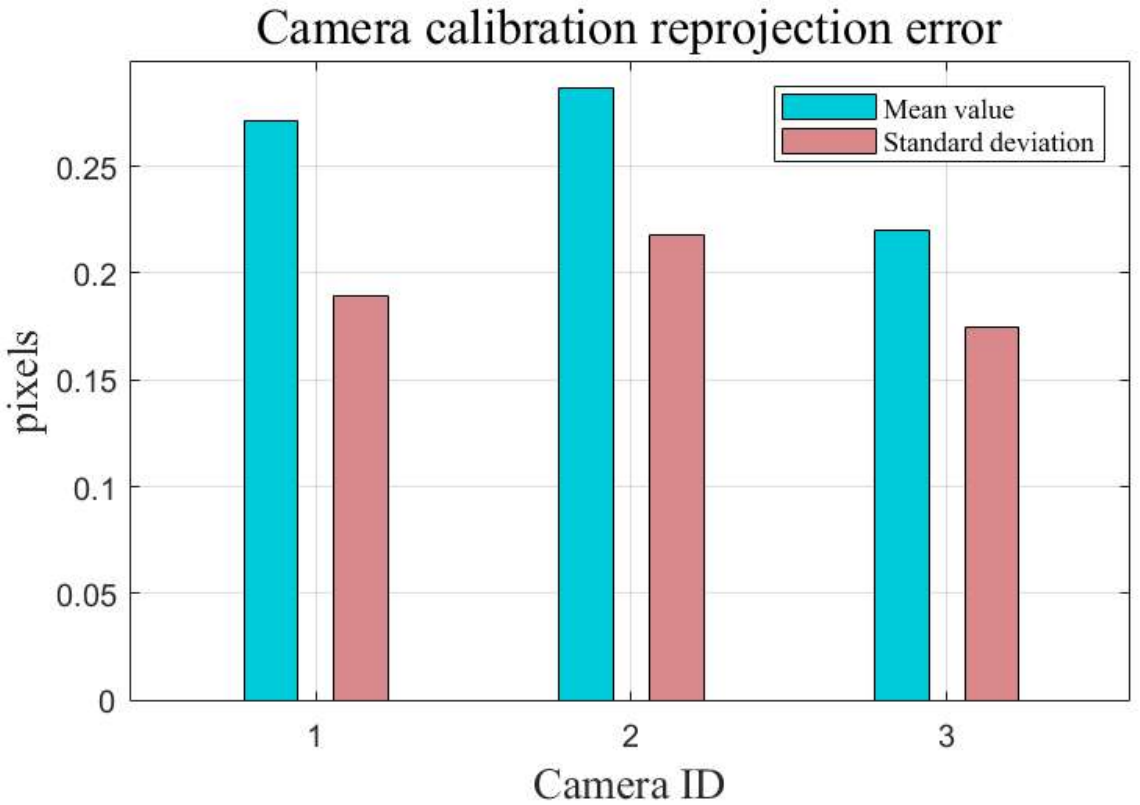}
    \caption{Mean and standard deviation of reprojection error for each event camera.}
    \label{fig5}
\end{figure}

The mean and standard deviation of the reprojection error for each event camera calibration are shown in Fig.~\ref{fig5}. The average reprojection error of event camera 1 is 0.27 pixels, with a standard deviation of 0.18 pixels, while the average reprojection error of event camera 2 is 0.28 pixels, with a standard deviation of 0.21 pixels. Additionally, the average reprojection error of event camera 3 is 0.22 pixels, with a standard deviation of 0.17 pixels. The reprojection error of the camera is less than 0.3 pixels.

\begin{figure}[htbp]
    \centering
    \includegraphics[width=0.4\linewidth]{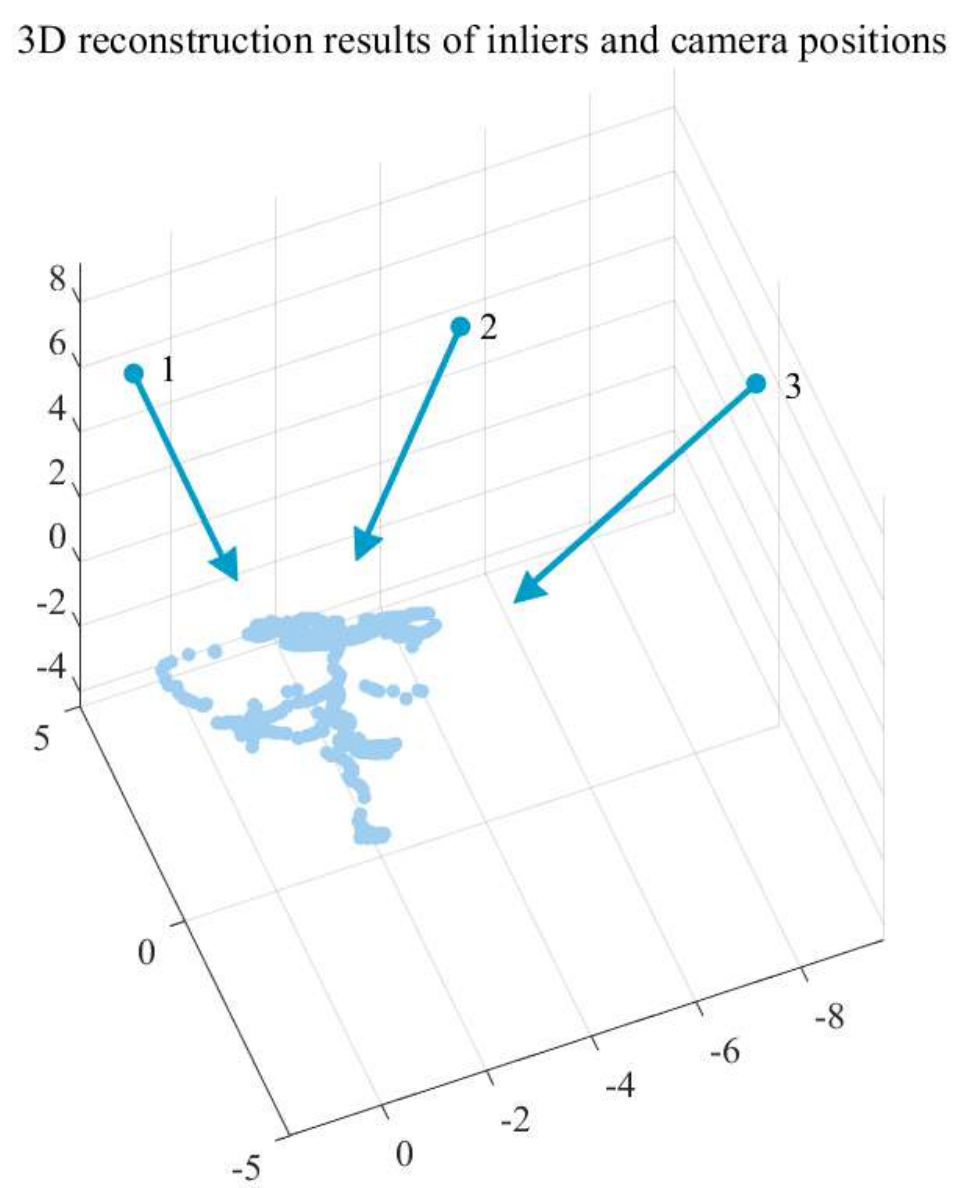}
    \caption{Reprojection error and 3D reconstruction results of inliers and camera positions.}
    \label{fig6}
\end{figure}

Fig.~\ref{fig6} shows the reconstructed 3D coordinates of the marker points after removing outliers based on the event camera's calibration parameters, as well as the event camera's position and optical axis direction. The results indicate that the optical axis direction of the relative position of the event camera is consistent with the actual experiment.

\begin{figure}[htbp]
    \centering
    \includegraphics[width=0.5\linewidth]{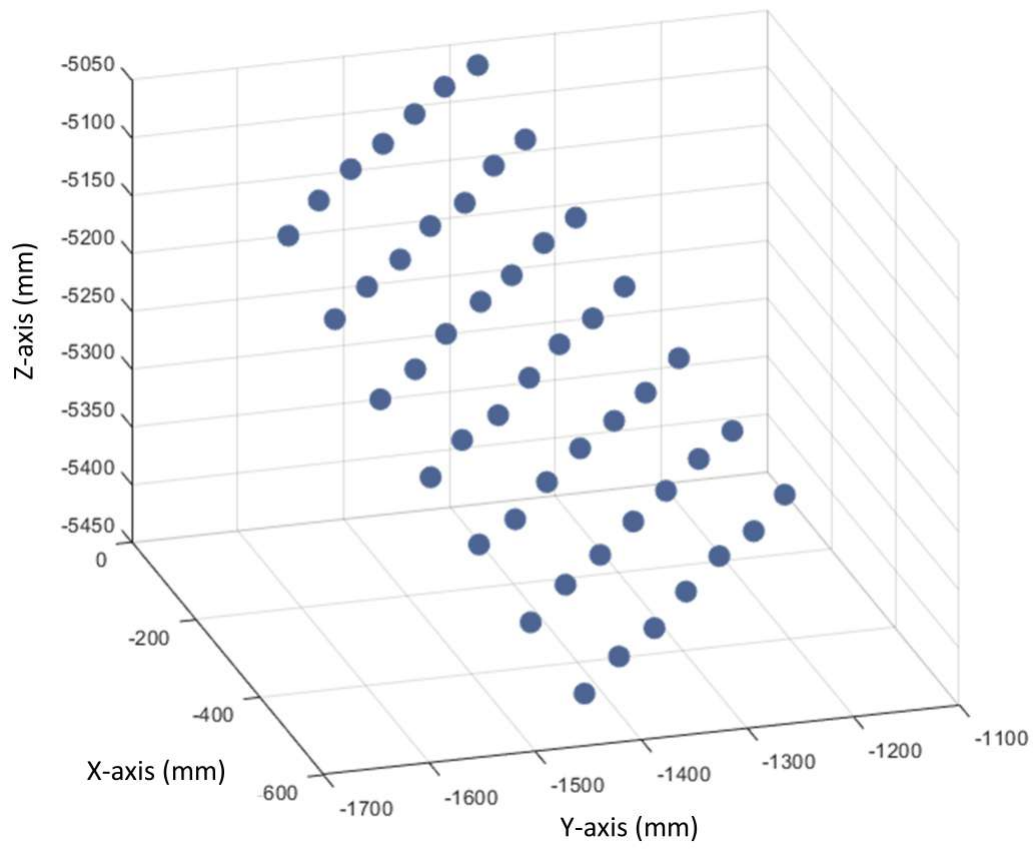}
    \caption{Actual distance measurement results of corner points.}
    \label{fig:new1}
\end{figure}

To validate the calibration accuracy of our method, we conduct the verification experiment. The actual baseline distance between cameras is used to scale the self‑calibration results. A checkerboard with a square size of 50mm was employed. The board was gently moved within the event camera’s field of view, allowing the camera to capture events triggered by the motion of the checkerboard edges. Based on the calibration parameters obtained by our method, the real‑world coordinates of each corner are reconstructed via spatial intersection, as is seen in Fig.~\ref{fig:new1}. The distances between selected corners are then measured and compared to their known physical dimensions. Because the edge features of the checkerboard observed by the event camera are not sharply defined, corner‑extraction errors significantly affect the measurements. To mitigate this, multiple sets of data are collected and averaged.

The actual measurement results are shown in Table~\ref{tabnew1} and Table~\ref{tabnew2}.

\begin{table*}[htbp]
\caption{Actual distance measurement results of camera 1 and camera 2.}\label{tabnew1}
    \centering
    \resizebox{\linewidth}{!}{
    \begin{tabular}{c c c c}
\toprule
Actual distance (mm) & Camera 1 and Camera 2 intersect (mm) & Absolute error (mm) & Relative error \\ 
\midrule
150 & 149.39 & 0.61 & 0.41\% \\ 
200 & 201.76 & 1.76 & 0.88\% \\ 
250 & 245.20 & 4.80 & 1.90\% \\ 
300 & 299.47 & 0.53 & 0.18\% \\ 
\botrule  
\end{tabular}}
\end{table*}

\begin{table*}[htbp]
\caption{Actual distance measurement results of camera 2 and camera 3.}\label{tabnew2}
    \centering
    \resizebox{\linewidth}{!}{
    \begin{tabular}{c c c c}
\toprule
Actual distance (mm) & Camera 1 and Camera 2 intersect (mm) & Absolute error (mm) & Relative error \\ 
\midrule
150 & 150.31 & 0.31 & 0.21\% \\ 
200 & 201.69 & 1.69 & 0.84\% \\ 
250 & 251.32 & 1.32 & 0.53\% \\ 
300 & 302.32 & 2.32 & 0.77\% \\ 
\botrule  
\end{tabular}}
\end{table*}

As shown in the table above, when measuring the distance between two adjacent corners, the result is more sensitive to extraction errors at each corner, leading to a larger relative error. As the measured distance between corners increases, the relative error decreases. For a corner spacing of 300mm, the relative difference between the triangulation results from camera 1 and camera 2 and those from camera 2 and camera 3 is less than 1\%. These experimental results confirm that the proposed self‑calibration method meets the required accuracy.

\subsection{High-Speed Deformation Measurement under Extreme Illumination Conditions}
\begin{figure}[ht]
\centering
\subfigure[Optical camera captures images.]{ 
\begin{minipage}{0.5\linewidth}
\centering 
\includegraphics[height = 4.5cm]{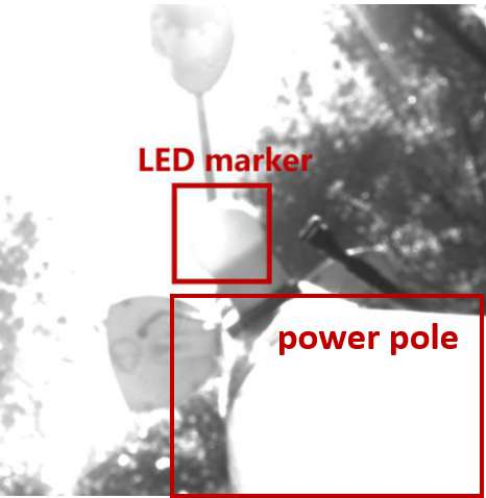}  
\end{minipage}
\label{fig7a}
}\subfigure[Event camera captures the event stream.]{ 
\begin{minipage}{0.5\linewidth}
\centering  
\includegraphics[height = 4.5cm]{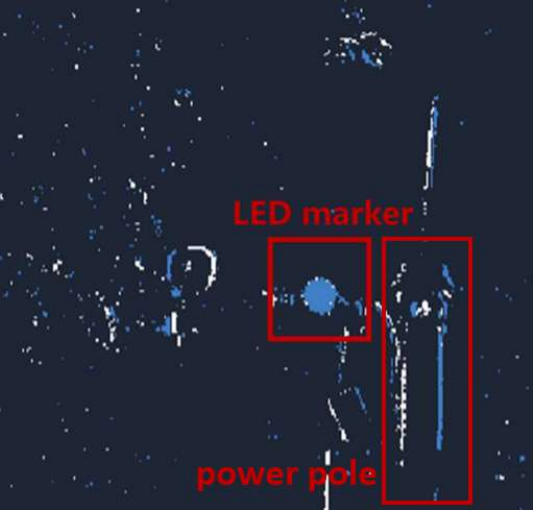}
\end{minipage}
\label{fig7b}
}
\caption{Imaging experiment under extreme illumination conditions.} 
\label{fig7}   
\end{figure}

To verify the performance of the proposed method for deformation measurement under extreme illumination conditions, the emulation experiment was first conducted. Under the same observational conditions, employ a traditional optical camera and an event camera to capture LED marker imagery independently. The result is shown in Fig.~\ref{fig7}. In Fig.~\ref{fig7a}, set the exposure time to 50 milliseconds and adjust the aperture to the minimum. At this point, there are still obvious overexposed areas in the image, making it impossible to extract marker features from the image. Accumulate the event stream into event frames at intervals of 1/30 s, as shown in Fig.~\ref{fig7b}. Clear symbol features are still visible in the event frames. This experiment demonstrates that the method proposed in this paper can accurately measure the 3D structural deformation under extreme illumination conditions, surpassing the capabilities of traditional cameras.

To validate measurement accuracy in the absence of known ground-truth structural deformation values, an emulation experiment was conducted. Two markers are mounted at both ends of a 1000.0 $\text{mm}$ pole, and the distance between markers is measured using the camera array presented in this paper. The LED marker was configured to blink at a frequency of 250 $\text{Hz}$. The marker center points were extracted by the proposed method, the measured maximum distance between the two markers was $999.2 \text{mm}$, yielding an absolute error of $0.8\text{mm}$ and a relative error of 0.08$\%$ compared to the truth value. The experimental results demonstrate that the proposed method achieves the required measurement accuracy.

Finally, we measured the high-speed deformation under extreme illumination conditions. Field environments exhibit highly heterogeneous illumination, characterized by alternating direct sunlight and shadow. Mount the LED marker on a tower positioned centrally within the camera array's field of view. Operate the LED marker at 250 Hz with synchronized pulse-width modulation, while applying mechanical excitation to the tower via sustained rapid oscillation. The multi-event camera array then quantifies the 3D displacement of LED markers using high-speed photogrammetry.
\begin{figure*}[htbp]
\centering
\subfigure[3D deformation measurement.]{ 
\begin{minipage}{0.5\linewidth}
\centering 
\includegraphics[width=0.9\linewidth]{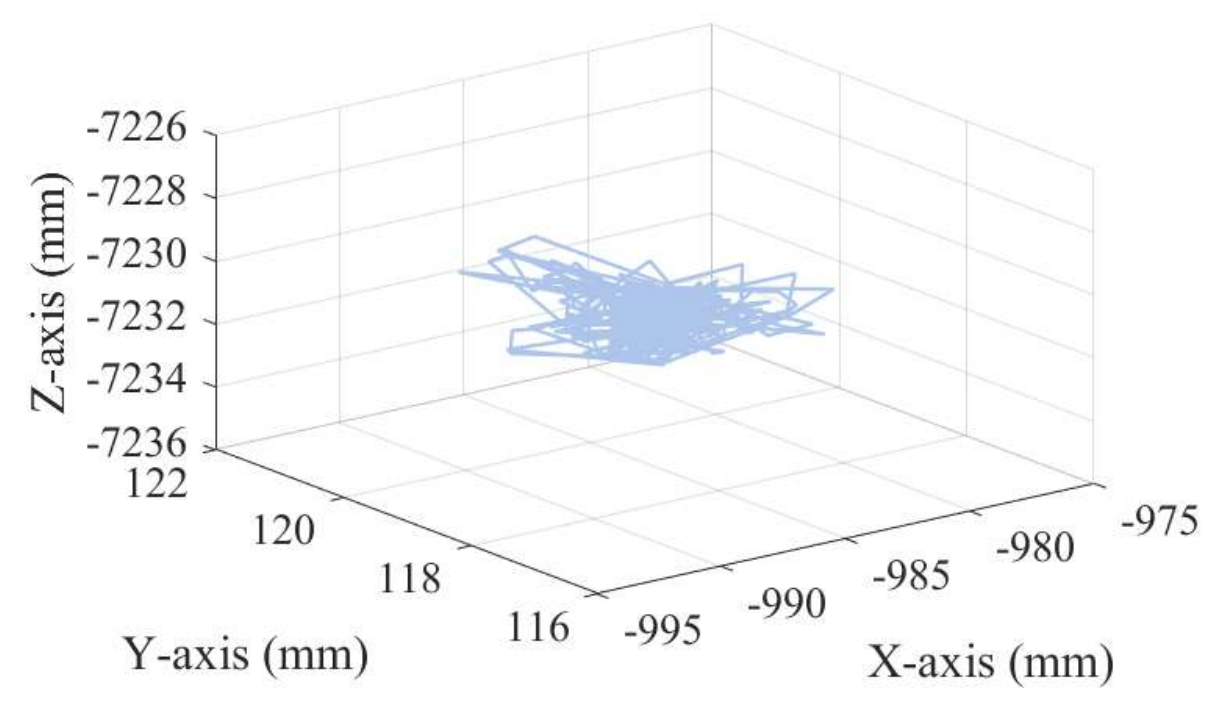}  
\end{minipage}
\label{fig9a}
}\subfigure[Component in the X-axis.]{ 
\begin{minipage}{0.5\linewidth}
\centering  
\includegraphics[width=1\linewidth]{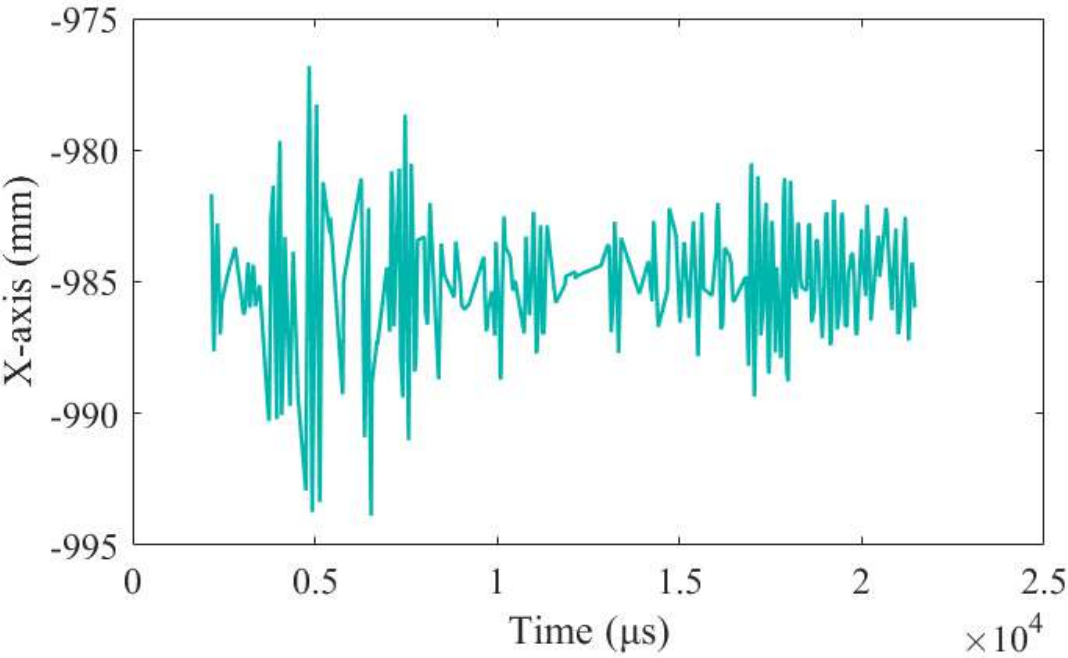}
\end{minipage}
\label{fig9b}
}
\subfigure[Component in the Y-axis.]{ 
\begin{minipage}{0.5\linewidth}
\centering  
\includegraphics[width=1\linewidth]{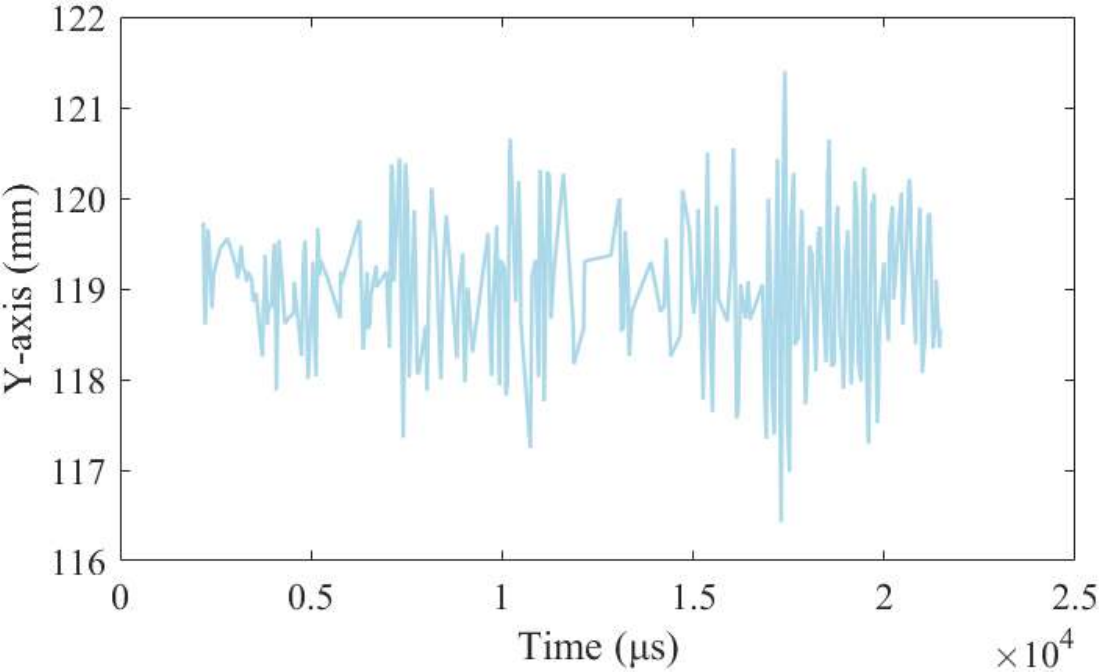}
\end{minipage}
\label{fig9c}
}\subfigure[Component in the Z-axis.]{ 
\begin{minipage}{0.5\linewidth}
\centering  
\includegraphics[width=1\linewidth]{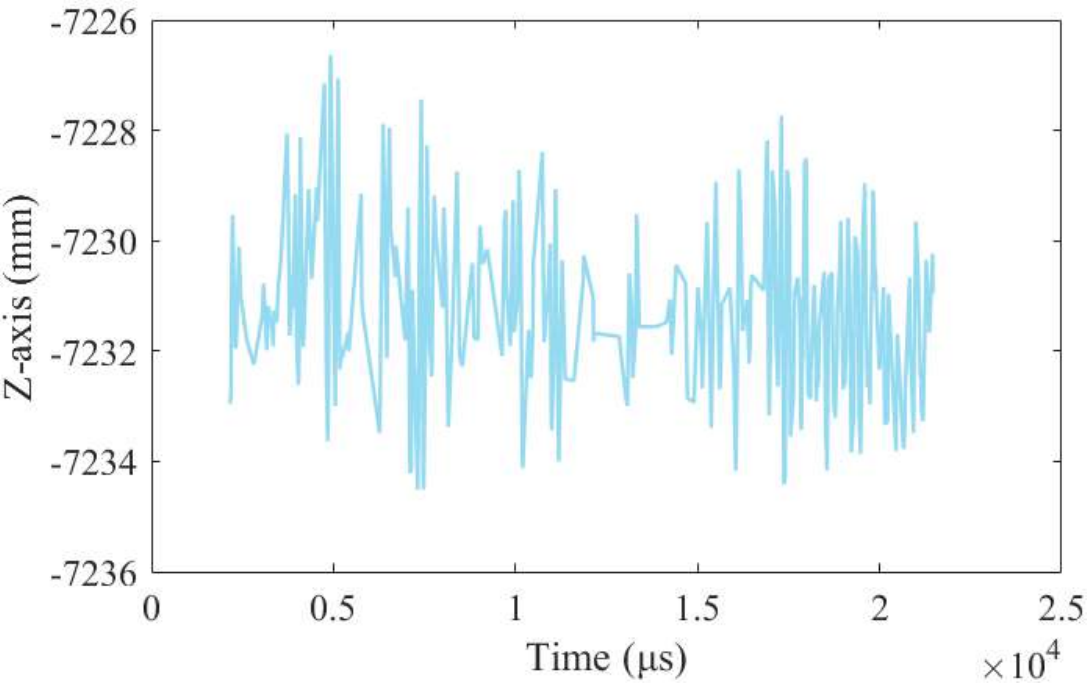}
\end{minipage}
\label{fig9d}
}
\caption{Deformation measurement results.} 
\label{fig9}   
\end{figure*}

3D deformation measurements of the tower structure were obtained through spatial intersection using the multi-event camera array, with results shown in Fig.~\ref{fig9}. The 3D deformation trajectory is seen in Fig.~\ref{fig9a}, which exhibits a maximum amplitude of 18.2 mm. Fig.~\ref{fig9b} shows the projection of the deformation in the X-axis direction, while Fig.~\ref{fig9c} and Fig.~\ref{fig9d} show the projections of the deformation in the Y-axis and Z-axis directions, respectively. The method described in this paper can accurately track the markers and achieve high-speed 3D deformation measurements of structures under extreme illumination conditions. Through the experiments mentioned above, the effectiveness of the entire methodology and workflow of the method, leveraging a synchronized multi-camera array, has been validated.

\section{Conclusion}\label{sec4}
This paper addresses the critical limitation of conventional cameras in measuring high-speed 3D deformation under extreme illumination. We present an integrated calibration to measurement method based on a multi-event camera array for monitoring structural deformation under extreme illumination conditions. The method first extracts corresponding marker centers by leveraging the asynchronous characteristics of the event stream combined with temporal correlation analysis. Camera calibration is then achieved efficiently through Kruppa equation solving integrated with a parameter optimization framework. Finally, 3D deformation under extreme lighting is reconstructed via unified coordinate transformation and linear intersection. Experimental results confirm that our method has advantages in accurately measuring 3D structural deformation under harsh illumination. The method offers a flexible and practical solution for field-based structural monitoring in challenging outdoor lighting. An inherent limitation of self-calibration is its inability to recover the absolute scale. To resolve this in practice, imposing an external metric constraint, such as a known baseline distance, is necessary.

\section*{Acknowledgements}
This research has been supported by the Hunan Provincial Natural Science Foundation for Excellent Young Scholars under Grant 2023JJ20045, and the National Natural Science Foundation of China under Grant 12372189.

\section*{Declarations}
\noindent $\textbf{Conflict of Interests}$ The authors declare that they have no known competing financial interests or personal relationships that could have appeared to influence the work reported in this paper.

\noindent $\textbf{Data Availability}$ Data underlying the results presented in this paper are not publicly available at this time but may be obtained from the authors upon reasonable request.

\section*{CRediT authorship contribution statement}
\noindent $\textbf{Banglei Guan}$: Investigation, Supervision, Methodology, Writing – review \& editing. 

\noindent $\textbf{Yifei Bian}$: Investigation, Software, Methodology, Writing – original draft. 

\noindent $\textbf{Zibin Liu}$: Investigation, Software. 

\noindent $\textbf{Haoyang Li}$: Investigation, Formal analysis.

\noindent $\textbf{Xuanyu Bai}$: Investigation, Validation.

\noindent $\textbf{Taihang Lei}$: Investigation, Formal analysis.

\noindent $\textbf{Bin Li}$: Investigation, Formal analysis.

\noindent $\textbf{Yang Shang}$: Supervision, Methodology, Writing –review \& editing. 

\noindent $\textbf{Qifeng Yu}$: Supervision, Methodology.

\bibliography{bibliography}
\end{document}